\documentclass[letterpaper]{article} 
\usepackage{aaai2026}  
\usepackage{times}  
\usepackage{helvet}  
\usepackage{courier}  
\usepackage[hyphens]{url}  
\usepackage{graphicx} 
\usepackage{natbib}  
\usepackage{caption} 
\usepackage{algorithm}
\usepackage{algorithmic}
\usepackage{amssymb}
\usepackage{amsmath}
\usepackage[T1]{fontenc}

\usepackage{multirow}
\usepackage{xcolor}
\usepackage{cleveref}

\usepackage{subfigure}

\usepackage{newfloat}
\usepackage{listings}
\DeclareCaptionStyle{ruled}{labelfont=normalfont,labelsep=colon,strut=off} 
\floatstyle{ruled}
\newfloat{listing}{tb}{lst}{}
\floatname{listing}{Listing}
\nocopyright 

\title{Neighbor-aware Instance Refining with Noisy Labels for Cross-Modal Retrieval}
\author{
    Yizhi Liu\textsuperscript{1,2}\thanks{Equal contribution.},
    Ruitao Pu\textsuperscript{1,2}\footnotemark[1],
    Shilin Xu\textsuperscript{1},
    Yingke Chen\textsuperscript{3},
    Quan-Hui Liu\textsuperscript{1},
    Yuan Sun \textsuperscript{1,2,4}\thanks{Corresponding author.}
}
\affiliations{
    \textsuperscript{\rm 1}College of Computer Science, Sichuan University, Chengdu, China, 610065\\
    \textsuperscript{\rm 2}State Key Laboratory of Al Safety, Beijing, China, 100086\\
    \textsuperscript{\rm 3}Department of Computer and Information Sciences, Northumbria University, Newcastle upon Tyne, UK, NE1 8ST\\
    \textsuperscript{\rm 4}National Key Laboratory of Fundamental Algorithms and Models for Engineering Numerical Simulation, Sichuan University, Chengdu, China, 610207\\
    liuyizhi@stu.scu.edu.cn, ruitaopu@gmail.com, xushilin990@gmail.com, yke.chen@gmail.com, quanhuiliu@scu.edu.cn, sunyuan\_work@163.com.

}

\usepackage{bibentry}

\begin{document}

\maketitle

\begin{abstract}
In recent years, Cross-Modal Retrieval (CMR) has made significant progress in the field of multi-modal analysis. However, since it is time-consuming and labor-intensive to collect large-scale and well-annotated data, the annotation of multi-modal data inevitably contains some noise. This will degrade the retrieval performance of the model. To tackle the problem, numerous robust CMR methods have been developed, including robust learning paradigms, label calibration strategies, and instance selection mechanisms. Unfortunately, they often fail to simultaneously satisfy model performance ceilings, calibration reliability, and data utilization rate. To overcome the limitations, we propose a novel robust cross-modal learning framework, namely Neighbor-aware Instance Refining with Noisy Labels (NIRNL). Specifically, we first propose Cross-modal Margin Preserving (CMP) to adjust the relative distance between positive and negative pairs, thereby enhancing the discrimination between sample pairs. Then, we propose Neighbor-aware Instance Refining (NIR) to identify pure subset, hard subset, and noisy subset through cross-modal neighborhood consensus. Afterward, we construct different tailored optimization strategies for this fine-grained partitioning, thereby maximizing the utilization of all available data while mitigating error propagation. Extensive experiments on three benchmark datasets demonstrate that NIRNL achieves state-of-the-art performance, exhibiting remarkable robustness, especially under high noise rates.


\end{abstract}

\begin{links}
    \link{Code}{https://github.com/perquisite/NIRNL}
\end{links}

\section{Introduction}
With the rapid development of multimodal data on the Internet, cross-modal retrieval (CMR) has become a research hot topic in the field of multimodal learning. Recently, a large number of CMR methods have been proposed, which aim to retrieve semantically related samples across heterogeneous modalities, such as image-text \cite{wehrmann2020adaptive,ge2023cross} or video-image scenarios~\cite{gorti2022x,fang2023uatvr}. Although these methods obtain the desired performance, most of them~\cite{zhen2019deep,pushe} rely heavily on clean-annotated data to learn multi-modal representations in a shared semantic space. In practice, collecting perfectly labeled data is both expensive and time-consuming. Due to annotation mistakes and ambiguous semantics, this inevitably results in noisy labels. The presence of noisy labels can severely harm the learning model, thereby weakening retrieval performance. To this end, some weakly supervised or semi-supervised CMR methods~\cite{mandal2019semi} have been proposed, which attempt to alleviate the influence of perfect labeling. However, they still have an implicit assumption, that is, the available labels are entirely correct. Therefore, how to robustly learn from noisy data has become a key challenge.

\begin{figure*}[]
\centering
\includegraphics[width=0.9\textwidth]{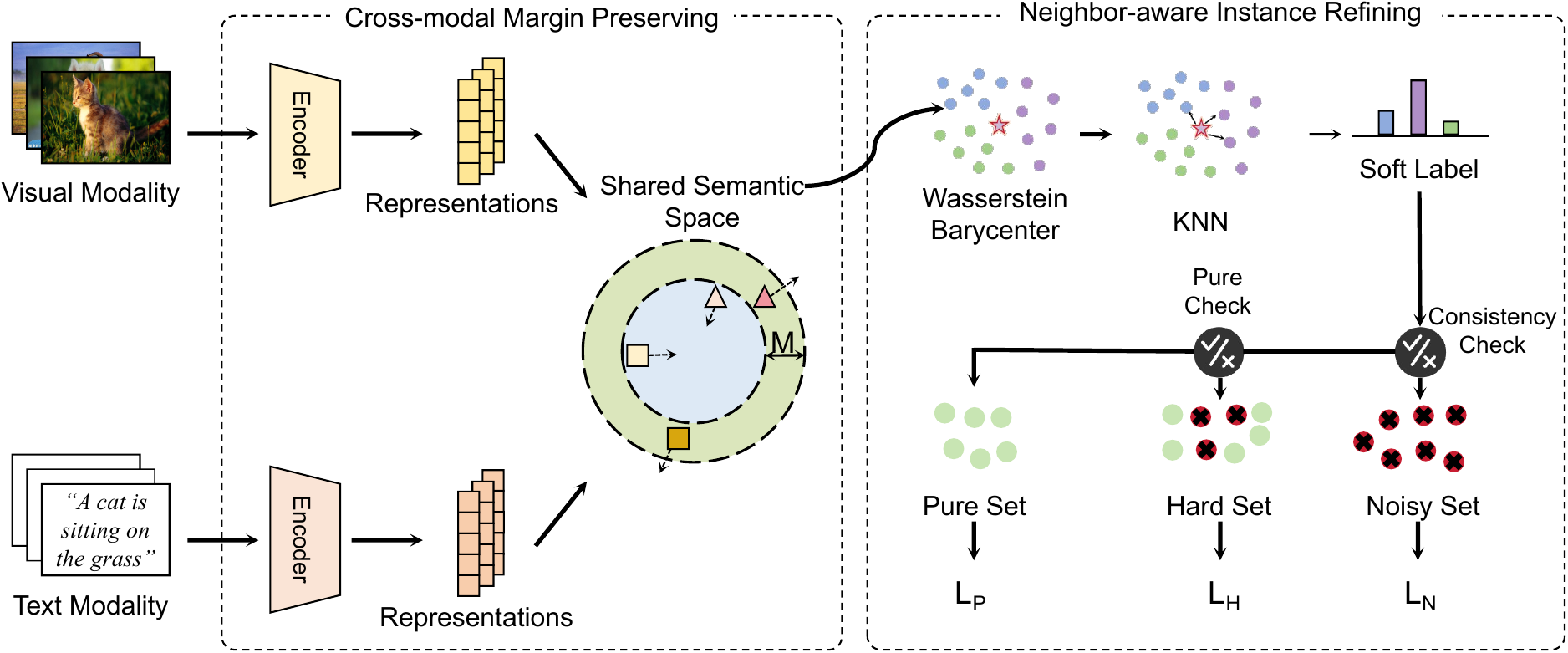}
\caption{The framework of our proposed NIRNL. Our framework comprises two core modules operating in parallel: \textbf{the Cross-modal Margin Preserving (CMP)} module and \textbf{the Neighbor-aware Instance Refining (NIR)} module. The CMP module refines the global structure of the embedding space, promoting proximity between positive pairs (indicated by light yellow and light pink) while enforcing separation of negative pairs (indicated by dark yellow and dark pink). For clarity, only image samples are visualized in the NIR module. The NIR module initially computes the Wasserstein Barycenter of samples and generates soft labels through KNN. It subsequently partitions the dataset into pure, hard, and noisy subsets by evaluating the consistency between soft labels and ground-truth labels. Finally, we design three different loss functions for each subset to dig up as much semantic information as possible.}
\label{fig:1}
\end{figure*}

Recently, various CMR methods \cite{fenginteractive, feng2023rono, pu2024deep, fengrobust, RSHNL, wang2024robust} have been proposed to robustly learn representations from multi-modal data with noisy labels. They could be roughly divided into three strategies, i.e., robust learning~\cite{hu2021learning, feng2023rono}, label calibration~\cite{pu2024deep, okamura2023lcnme}, and instance selection~\cite{RSHNL, wang2024robust}. To be specific, robust learning aims to design a robust loss, thereby directly tolerating the influence of noisy labels. To directly eliminate the influence of noisy labels, label calibration refines the labels from the source, thus improving the upper limit of model performance. To filter out instances with noisy labels, instance selection first identifies the wrong labels and then trains the model with clean data. Although they have demonstrated promising outcomes,  these strategies generally suffer from some limitations. For example, robust learning relies on prior assumptions about noise distribution and can only tolerate noise, but cannot eliminate its limitations on the upper limit of model performance. Label calibration could introduce new noise or amplify the errors when class boundaries are ambiguous or the noise distribution heavily overlaps with the true distribution. Instance selection is sensitive to the pre-set threshold, which can easily lead to the filtering of clean instances or the omission of noisy instances. Meanwhile, it could also result in a significant waste of training data. In summary, dynamically coordinating model performance ceilings, calibration reliability, and data utilization efficiency under complex noise scenarios remains a key yet challenging problem.

To overcome the aforementioned challenges, we propose a novel Neighbor-aware Instance Refining with Noisy Labels (NIRNL) framework to mitigate the negative effects of noisy labels. As shown in Fig.\ref{fig:1}, NIRNL is composed of two core modules, i.e., Cross-modal Margin Preserving (CMP) and Neighbor-aware Instance Refining (NIR). 
Firstly, CMP imposes constraints on the relative distances between positive and negative pairs to enhance the discriminability of representations in the shared semantic space.
Then, NIR retrieves nearest neighbors and evaluates their consistency with ground-truth labels to generate soft labels, which could effectively partition these instances into three distinct subsets,i.e., pure, hard, and noisy subsets.
For the pure subset, where labels are highly reliable, NIR optimizes instances directly to fully exploit their supervisory signals. For the hard subset, where label reliability is uncertain, NIR applies a weighted optimization strategy to balance the influence of potentially corrupted annotations while retaining useful information. For the noisy subset, where labels are deemed unreliable, NIR performs label calibration to recover informative content and mitigate the adverse effects of noise.
The main contributions are summarized as follows:
\begin{itemize}
\item To achieve high tolerance to noisy labels, we propose a robust cross-modal learning framework (NIRNL), which unifies robust learning, label calibration, and instance selection. To the best of our knowledge, this is the first work to balance model performance ceiling, calibration reliability, and data utilization rate in multi-modal learning with noisy labels.
\item We propose neighbor-aware instance refining that dynamically partitions training instances into pure, hard, and noisy subsets by perceiving the global neighborhood distribution. Further, we design a customized optimization strategy to explore all available information as much as possible, thereby enhancing the robustness.
\item Extensive experiments comprehensively verify that our proposed NIRNL has remarkably superior performance over the current state-of-the-art methods.


\end{itemize}

\section{Related Work}

\subsection{Cross-modal Retrieval}

With the exponential growth of internet data, cross-modal retrieval (CMR) has emerged as a key technology for information retrieval across diverse data types~\cite{su2025dica,li2025robust,li2024romo,li2025learning}. The fundamental challenge in CMR lies in bridging the heterogeneity gap between different modalities~\cite{ijcai2025p845,sun2023hierarchical,yin2025roda}. Some methods focus on unsupervised methods. Early approaches employed single methods, such as UDCMH~\cite{wu2018unsupervised}, which extract features and generate hash codes as pseudo-labels for learning. However, these methods heavily relied on the quality of the initial similarity matrix. DJSRH~\cite{su2019deep} utilizes matrix fusion to incorporate neighborhood information from all modalities, but this introduced redundancy. Further advancing the field, researchers have adopted hybrid models, such as UGACH~\cite{zhang2018unsupervised}, UCCH~\cite{hu2022unsupervised}, and UCGKANH~\cite{lin2025unsupervised}. However, the performance of these methods is inherently constrained by their reliance on pseudo-labels generated from the data itself.

Consequently, some supervised methods leverage explicit label information to learn a shared semantic space. These methods leverage uncertainty modeling. PCME~\cite{chun2021probabilistic} regards each sample as a probability distribution in the embedding space but heavily relies on paired data. Building upon this, DECL~\cite{qin2022deep} employs evidential learning to quantify uncertainty caused by potentially noisy correspondences, yet it lacks uncertainty modeling for individual outcomes. Moreover, some methods have explored adversarial learning~\cite{wang2017adversarial}, graph neural networks DAGN~\cite{qian2022integrating,liang2023knowledge}, consistency learning DRCL~\cite{pu2025deep}. Although the supervised methods discussed above have achieved impressive performance on various benchmarks, their success is largely predicated on a critical assumption: the availability of large-scale, meticulously annotated training datasets. 

\subsection{Cross-modal Retrieval with Noisy Labels}
In practice, however, collecting and annotating such high-quality data is both prohibitively expensive and highly time-consuming. Consequently, researchers often turn to collecting web data, which offers scale, but is noisy, misguides training, corrupts semantic alignment, and hurts retrieval~\cite{han2025unified}. Noise-robust learning is now critical for real-world cross-modal search. These methods generally fall into three main categories.

Robust Learning methods design loss functions or training strategies that tolerate noise. For example, RONO~\cite{feng2023rono} adopts discriminative center learning to pull clean samples closer and push noisy ones away. However, these methods suppress rather than correct noise.
Instance Selection methods filter out noisy samples and train on clean subsets. RSHNL~\cite{pu2025deep} uses a self-paced learning strategy, while NRCH~\cite{wang2024robust} applies the small-loss criterion. Yet, these methods often discard a large portion of data, including hard-but-clean samples that are essential for learning robust decision boundaries.
Label Calibration methods take a direct approach by correcting noisy labels. UOT-RCL formulates label correction as a partial Optimal Transport problem to align noisy labels with corrected ones, while TCL~\cite{li2021trustable} leverages a small trusted dataset to guide co-learning. Although effective, these methods risk introducing new errors or amplifying existing ones.

However, these methods struggle to strike a balance between model performance ceilings, calibration reliability, and data utilization efficiency. To address this, our proposed NIRNL method uses consensus signals from heterogeneous sources to identify, refine, and fully exploit the data.
\section{The Proposed Method}

\subsection{Preliminaries}
Without loss of generality, taking the visual-text retrieval as an example, we define some denotations to present the learning with noisy labels (LNL) problem in cross-modal retrieval (CMR). Considering a dataset with noisy labels $\mathcal{D} =\left \{\mathcal{V}_i,\mathcal{T}_i,\mathcal{Y}_i  \right \} _{i=1}^N$ with $N$ instances, where $(\mathcal{V}_i,\mathcal{T}_i)$ is the visual-text pair and $\mathcal{Y}_i = \left \{ \mathcal{Y}_i^1, \mathcal{Y}_i^2,...,\mathcal{Y}_i^C\right \}  \in\left \{ 0,1 \right \} ^{1 \times C}$ denotes the corresponding noisy label over $C$ categories. Further, if the $i$-th instance belongs to the $c$-th category, $\mathcal{Y}_i^c=1$, otherwise $\mathcal{Y}_i^c=0$. The fundamental goal of CMR is to learn a series of modality-specific sub-networks responsible for projecting multimodal data into the common semantic space where cross-modal similarity can be directly measured. Let the dimensionality of the common semantic space be $d$, sub-networks for the visual and text modalities are denoted as $\mathcal{F}^{\mathcal{V} } (\cdot,\Phi ^{\mathcal{V} })\in \mathbb{R}^{1 \times d}$ and $\mathcal{F}^{\mathcal{T} } (\cdot,\Phi ^{\mathcal{T} })\in \mathbb{R}^{1 \times d} $, respectively, where $\Phi ^{\mathcal{V}}$ and $\Phi ^{\mathcal{T} }$ are the learnable parameters. For brevity, we use $f_i^\mathcal{V}$ and $f_i^\mathcal{T}$ to their outputs, i.e.,
$f_i^\mathcal{V} = \mathcal{F}^{\mathcal{V} } (\mathcal{V}_i)$ and $f_i^\mathcal{T} = \mathcal{F}^{\mathcal{T} } (\mathcal{T}_i)$, respectively.


\subsection{Cross-modal Margin Preserving}
To strengthen the distinguishability among sample pairs, we propose Cross-modal Margin Preserving (CMP) to constrain the relative distance between positive and negative pairs, thereby yielding more compact representations for intra-class samples and differentiated representations for inter-class ones. Specifically, CMP enforces that the similarity of positive pairs exceeds that of negative pairs by at least a predefined margin. The CMP loss could be formulated as follows:
\begin{equation}
\small
\begin{aligned}
    \mathcal{L}_{CMP} &= \frac{1}{N} \sum_{i=1}^{N} \sum_{\substack{j=1 \\ j \ne i}}^{N} | \Gamma(f_i^{\mathcal{V}}, f_j^{\mathcal{T}}) - \Gamma(f_i^{\mathcal{V}}, f_i^{\mathcal{T}}) + \mathcal{M} |_+ \\
&+ \frac{1}{N} \sum_{i=1}^{N} \sum_{\substack{j=1 \\ j \ne i}}^{N}| \Gamma(f_i^{\mathcal{T}}, f_j^{\mathcal{V}}) - \Gamma(f_i^{\mathcal{T}}, f_i^{\mathcal{V}}) + \mathcal{M} |_+,
\end{aligned}
\label{eq:12}
\end{equation}
where $\mathcal{M}$ is the margin and $|\cdot|_+$ denotes the hinge function, which returns the input if it is positive and zero otherwise, i.e., $|x|_+ = \max(0, x)$.

\subsection{Neighbor-aware Instance Refining}
To mitigate the negative impact of noisy labels, previous methods \cite{yang2022mutual, RSHNL} typically select confident instances in small batches based on the small-loss criterion to guide model training. However, this strategy overlooks the global neighborhood structure among samples and discards potentially informative instances, making the performance of the model susceptible to the training data distribution and prone to performance bottlenecks. For this issue, we propose Neighbor-aware Instance Refining (NIR) to integrate neighborhood consistency for generating soft labels, thereby achieving a refined partitioning of samples and formulating corresponding learning strategies to maximize the use of all available information. First, NIR generates soft labels for the two modalities by using, 
\begin{equation}
\small
\begin{aligned}
    \hat{p}(c \mid \mathcal{V}_i) & = \frac{1}{K} \sum_{k=1, \mathcal{V}_k \in \mathcal{N}_i^{\mathcal{V}}}^{K} \mathbb{I}[y_k ^ c = 1], \quad c \in [1, 2, \dots, C], \\
    \hat{p}(c \mid \mathcal{T}_i) & = \frac{1}{K} \sum_{k=1, \mathcal{T}_k \in \mathcal{N}_i^{\mathcal{T}}}^{K} \mathbb{I}[y_k ^ c = 1], \quad c \in [1, 2, \dots, C],
\end{aligned}
\label{eq:1}
\end{equation}
where $\mathcal{N}_i^{\mathcal{V}}$ and $\mathcal{N}_i^{\mathcal{T}}$ denote the sets of the $K$ nearest neighbors of $\mathcal{V}_i$ and $\mathcal{T}_i$ in the visual and textual modalities, respectively. Then, we evaluate the consistency between the learned soft labels and ground-truth labels to divide instances into three subsets, i.e., pure subset, hard subset, and noisy subset. Among these, if an instance achieves consistency in two modalities, and it can be incorporated into the pure subset, which could be written as:
\begin{equation}
\begin{aligned}
    \mathcal{D}_P=& \{ (\mathcal{V}_i,\mathcal{T}_i,\mathcal{Y}_i): \mathcal{Y}_i^{\operatorname{argmax}\limits_{c}  \hat{p}(c \mid \mathcal{V}_i)} =1 \, \And \\ &\mathcal{Y}_i^{\operatorname{argmax}\limits_{c}  \hat{p}(c \mid \mathcal{T}_i)} =1, i \in \left [  1,2,...,N \right ]  \} .
\end{aligned}
\label{eq:2}
\end{equation}
Further, if an instance only achieves consistency in one modality, and it can be incorporated into the hard subset, which could be formulated as:
\begin{equation}
\begin{aligned}
    \mathcal{D}_H=& \{ (\mathcal{V}_i,\mathcal{T}_i,\mathcal{Y}_i): \mathcal{Y}_i^{\operatorname{argmax}\limits_{c}  \hat{p}(c \mid \mathcal{V}_i)} =1 \, | \\ &\mathcal{Y}_i^{\operatorname{argmax}\limits_{c}  \hat{p}(c \mid \mathcal{T}_i)} =1, i \in \left [  1,2,...,N \right ]   \}-\mathcal{D}_P  .
\end{aligned}
\label{eq:3}
\end{equation}
At last, if an instance can not achieve consistency in any modalities, and it can be incorporated into the noisy subset, which could be given by:
\begin{equation}
\begin{aligned}
    \mathcal{D}_N=& \{ (\mathcal{V}_i,\mathcal{T}_i,\mathcal{Y}_i): \mathcal{Y}_i^{\operatorname{argmax}\limits_{c}  \hat{p}(c \mid \mathcal{V}_i)} =0 \, \And \\ &\mathcal{Y}_i^{\operatorname{argmax}\limits_{c}  \hat{p}(c \mid \mathcal{T}_i)} =0, i \in \left [  1,2,...,N \right ]  \} .
\end{aligned}
\label{eq:4}
\end{equation}
For the three subsets, we deploy corresponding optimization strategies to learn semantic-consistency representations under the interference of noisy labels. To achieve this, we first extract semantic barycenters \cite{agueh2011barycenters} for each class based on the feature distribution, and this process could be expressed as:
\begin{equation}
\small
\begin{aligned}
    &\bar{u}_c = \arg\min_{u} \sum_{i=1,\mathcal{Y}_i^c =1}^{N_c} \sum_{* \in \{\mathcal{V}, \mathcal{T}\}} \omega_i^{*} \cdot \mathcal{W}_{2,\lambda}^2(u, f_i^{*}), \\
    &\text{s.t.} \;\sum_{i=1,\mathcal{Y}_i^c =1}^{N_c} \sum_{* \in \{\mathcal{V}, \mathcal{T}\}} \omega_i^{*} = 1,  \forall \, \omega_i^{*} \ge 0, c \in [0,1,...,C],
\end{aligned}
\label{eq:5}
\end{equation}
where $N_c$ is the number of samples belonging to class $c$, $\omega_j^{*}$ denotes the weight for each sample (uniform weight in this paper), $\mathcal{W}_2^2(\cdot,\cdot)$ denotes the 2-Wasserstein distance, and $\lambda$ is the regularization coefficient. In our paper, we use the expectation maximization (EM) algorithm \cite{dempster1977maximum, feng2023ot} to converge the barycenters. 

After mining barycenters, for the $i$-th sample in two modalities, we can obtain its probability that it belongs to its corresponding barycenter as follows:
\begin{equation}
    \begin{aligned}
        s(*_i)=\sum_{c=1}^{C} \mathcal{Y}_i^c\frac{\Gamma(f_i^*, \bar{u}_c)}{\sum_{j=1}^{C} \Gamma(f_i^*, \bar{u}_j)}, * \in \{\mathcal{V}, \mathcal{T}\}, 
    \end{aligned}
    \label{eq:6}
\end{equation}
where $\gamma(\cdot,\cdot)$ refers to the cosine similarity operator. Then, we adopt different optimization techniques for $\mathcal{D}_P$, $\mathcal{D}_H$, and $\mathcal{D}_N$. For the pure subset $\mathcal{D}_P$, we assume with high confidence that the labels are reliable and expect each sample to align closely with the corresponding semantic barycenter. To this end, we adopt the cross-entropy (CE) loss, which is known for its strong generalization ability \cite{ghosh2017robust} to optimize the model, i.e.,
\begin{equation}
\begin{aligned}
\mathcal{L}_P = -\frac{1}{|\mathcal{D}_P|} \sum_{i=1}^{|\mathcal{D}_P|} \sum_{* \in \{\mathcal{V}, \mathcal{T}\}} \log  s(*_i),
\end{aligned}
    \label{eq:7}
\end{equation}
where $|\cdot|$ means the size of a set. For the hard subset $\mathcal{D}_H$, we believe that a majority of the samples are still clean, but due to label uncertainty, we cannot be fully confident. To address this, we adopt a loss-weighting strategy that allows the model to place greater emphasis on instances that are more likely to be clean, thereby improving robustness. The weight for each instance can be obtained as follows:
\begin{equation}
        \ell_i=1-(1-s(\mathcal{V}_i ))\cdot (1-s(\mathcal{T}_i )).
    \label{eq:8}
\end{equation}
Then, we guide the model optimization by adopting the weighted CE loss as follows:
\begin{equation}
\mathcal{L}_H = -\frac{1}{|\mathcal{D}_H|} \sum_{i=1}^{|\mathcal{D}_H|} \ell_i\sum_{* \in \{\mathcal{V}, \mathcal{T}\}} \log  s(*_i).
    \label{eq:9}
\end{equation}
For the noisy subset $\mathcal{D}_N$, we assume with high confidence that all samples are mislabeled and apply a label correction strategy to exploit useful information from them. For the $i$-th instance, we first fuse the soft labels from two modalities as follows:
\begin{equation}
    \hat{p}_i ^c= 1 - (1- \hat{p}(c|\mathcal{V}_i ))\cdot (1- \hat{p}(c|\mathcal{T}_i )), \,c \in [1,2,...,C]
    \label{eq:10}
\end{equation}
Then, we use the maximum value of $\hat{p}_i ^c$ as the corrected label, i.e., $\hat{y}_i = \operatorname{argmax}\limits_{c} \hat{p}_i ^c$. Further, we construct the corresponding one-hot encoded label  $\hat{\mathcal{Y}}_i \in \{0,1\}^C$, where $\mathcal{Y}_i^c = 1$ if and only if $c = \hat{y}_i$. At last, recognizing that label correction may introduce biased predictions in complex data distributions, we mitigate the risk of overfitting caused by accumulated historical errors by the robust Mean Absolute Error (MAE) loss \cite{ghosh2017robust} as follows:
\begin{equation}
\small
\mathcal{L}_N = -\frac{1}{|\mathcal{D}_N|} \sum_{i=1}^{|\mathcal{D}_N|} \sum_{* \in \{\mathcal{V}, \mathcal{T}\}} \left ( 1-\frac{\Gamma(f_i^*, \bar{u}_{\hat{y}_i })}{\sum_{j=1}^{C} \Gamma(f_i^*, \bar{u}_j)} \right ).
    \label{eq:11}
\end{equation}



\subsection{Optimization}
Combining the aforementioned loss functions, the objective loss function of NIRNL is shown as follows:
\begin{equation}
    \mathcal{L} = (\mathcal{L}_P + \mathcal{L}_H + \mathcal{L}_N )+ \alpha \mathcal{L}_{CMP},
    \label{eq:13}
\end{equation}
where $\alpha$ is the balance coefficient to control the contribution of the corresponding loss term. Notably, the first three terms are applied only to samples within their corresponding subsets (i.e., $\mathcal{D}_P$, $\mathcal{D}_H$, and $\mathcal{D}_N$), while the last term is applied to all samples.

\section{Experiments}

\begin{table}[h]
\centering
\setlength{\tabcolsep}{1pt}
\begin{tabular}{c|ccccccccc}
\hline
\multirow{3}{*}{Method} & \multicolumn{9}{c}{Wikipedia} \\ \cline{2-10} 
& \multicolumn{2}{c|}{0.2} & \multicolumn{2}{c|}{0.4} & \multicolumn{2}{c|}{0.6} & \multicolumn{2}{c|}{0.8} & \multirow{2}{*}{Mean} \\ 
& I2T & \multicolumn{1}{c|}{T2I} & I2T & \multicolumn{1}{c|}{T2I} & I2T & \multicolumn{1}{c|}{T2I} & I2T & \multicolumn{1}{c|}{T2I} &  \\ \hline
DGCPN & 31.5 & \multicolumn{1}{c|}{26.9} & 31.5 & \multicolumn{1}{c|}{26.9} & 31.5 & \multicolumn{1}{c|}{26.9} & 31.5 & \multicolumn{1}{c|}{26.9} & 29.2 \\
CIRH & 23.6 & \multicolumn{1}{c|}{22.5} & 23.6 & \multicolumn{1}{c|}{22.5} & 23.6 & \multicolumn{1}{c|}{22.5} & 23.6 & \multicolumn{1}{c|}{22.5} & 23.1 \\
UCCH & 38.6 & \multicolumn{1}{c|}{37.4} & 38.6 & \multicolumn{1}{c|}{37.4} & 38.6 & \multicolumn{1}{c|}{37.4} & 38.6 & \multicolumn{1}{c|}{37.4} & 38.0 \\
ALGCN & 41.9 & \multicolumn{1}{c|}{40.1} & 30.9 & \multicolumn{1}{c|}{28.8} & 18.4 & \multicolumn{1}{c|}{17.2} & 13.4 & \multicolumn{1}{c|}{14.1} & 25.6 \\
RONO & 50.5 & \multicolumn{1}{c|}{47.1} & 48.8 & \multicolumn{1}{c|}{45.8} & 45.3 & \multicolumn{1}{c|}{41.8} & 41.6 & \multicolumn{1}{c|}{38.2} & 44.9 \\
GNN4CMR & 47.6 & \multicolumn{1}{c|}{44.1} & 42.0 & \multicolumn{1}{c|}{39.8} & 32.3 & \multicolumn{1}{c|}{30.6} & 20.9 & \multicolumn{1}{c|}{20.4} & 34.7 \\
DHRL & 46.2 & \multicolumn{1}{c|}{42.8} & 45.0 & \multicolumn{1}{c|}{42.2} & 35.5 & \multicolumn{1}{c|}{33.3} & 24.0 & \multicolumn{1}{c|}{24.5} & 36.7 \\
NRCH & 43.6 & \multicolumn{1}{c|}{42.3} & 39.5 & \multicolumn{1}{c|}{37.7} & 33.6 & \multicolumn{1}{c|}{32.0} & 31.6 & \multicolumn{1}{c|}{31.2} & 36.4 \\
DRCL & 47.5 & \multicolumn{1}{c|}{41.6} & 36.5 & \multicolumn{1}{c|}{32.3} & 23.4 & \multicolumn{1}{c|}{20.4} & 16.1 & \multicolumn{1}{c|}{14.7} & 29.1 \\
RSHNL & \underline{49.1} & \multicolumn{1}{c|}{\underline{45.4}} & \underline{44.3} & \multicolumn{1}{c|}{\underline{41.6}} & \underline{38.3} & \multicolumn{1}{c|}{\underline{36.4}} & \underline{27.8} & \multicolumn{1}{c|}{\underline{26.8}} & \underline{38.7} \\
NIRNL & \textbf{51.6} & \multicolumn{1}{c|}{\textbf{46.6}} & \textbf{51.7} & \multicolumn{1}{c|}{\textbf{46.5}} & \textbf{49.2} & \multicolumn{1}{c|}{\textbf{46.1}} & \textbf{41.7} & \multicolumn{1}{c|}{\textbf{39.4}} & \textbf{46.6} \\ \hline
\end{tabular}
\caption{The MAP scores with different noise rates.}
\label{tab:1}
\end{table}

\begin{table*}[h]
\centering
\setlength{\tabcolsep}{1pt}
\begin{tabular}{c|c|ccccccccc|ccccccccc}
\hline
\multirow{3}{*}{Method} & \multirow{3}{*}{Ref.} & \multicolumn{9}{c|}{XMedia} & \multicolumn{9}{c}{INRIA-Websearch} \\ \cline{3-20} 
 &  & \multicolumn{2}{c|}{0.2} & \multicolumn{2}{c|}{0.4} & \multicolumn{2}{c|}{0.6} & \multicolumn{2}{c|}{0.8} & \multirow{2}{*}{Mean} & \multicolumn{2}{c|}{0.2} & \multicolumn{2}{c|}{0.4} & \multicolumn{2}{c|}{0.6} & \multicolumn{2}{c|}{0.8} & \multirow{2}{*}{Mean} \\ 
 &  & I2T & \multicolumn{1}{c|}{T2I} & I2T & \multicolumn{1}{c|}{T2I} & I2T & \multicolumn{1}{c|}{T2I} & I2T & \multicolumn{1}{c|}{T2I} &  & I2T & \multicolumn{1}{c|}{T2I} & I2T & \multicolumn{1}{c|}{T2I} & I2T & \multicolumn{1}{c|}{T2I} & I2T & \multicolumn{1}{c|}{T2I} &  \\ \hline
DGCPN & AAAI'21 & 48.4 & \multicolumn{1}{c|}{33.6} & 48.4 & \multicolumn{1}{c|}{33.6} & 48.4 & \multicolumn{1}{c|}{33.6} & 48.4 & \multicolumn{1}{c|}{33.6} & 41.0 & 4.6 & \multicolumn{1}{c|}{4.4} & 4.6 & \multicolumn{1}{c|}{4.4} & 4.6 & \multicolumn{1}{c|}{4.4} & 4.6 & \multicolumn{1}{c|}{4.4} & 4.5 \\
CIRH & TKDE'22 & 83.9 & \multicolumn{1}{c|}{82.0} & 83.9 & \multicolumn{1}{c|}{82.0} & 83.9 & \multicolumn{1}{c|}{82.0} & 83.9 & \multicolumn{1}{c|}{82.0} & 83.0 & 30.9 & \multicolumn{1}{c|}{31.2} & 30.9 & \multicolumn{1}{c|}{31.2} & 30.9 & \multicolumn{1}{c|}{31.2} & 30.9 & \multicolumn{1}{c|}{31.2} & 31.1 \\
UCCH & TPAMI'23 & 78.6 & \multicolumn{1}{c|}{79.5} & 78.6 & \multicolumn{1}{c|}{79.5} & 78.6 & \multicolumn{1}{c|}{79.5} & 78.6 & \multicolumn{1}{c|}{79.5} & 79.1 & 7.4 & \multicolumn{1}{c|}{6.2} & 7.4 & \multicolumn{1}{c|}{6.2} & 7.4 & \multicolumn{1}{c|}{6.2} & 7.4 & \multicolumn{1}{c|}{6.2} & 6.8 \\
ALGCN & TMM'21 & 85.6 & \multicolumn{1}{c|}{84.6} & 74.7 & \multicolumn{1}{c|}{72.3} & 44.3 & \multicolumn{1}{c|}{40.2} & 14.1 & \multicolumn{1}{c|}{12.1} & 53.5 & 29.6 & \multicolumn{1}{c|}{28.1} & 17.4 & \multicolumn{1}{c|}{16.0} & 6.7 & \multicolumn{1}{c|}{6.2} & 2.4 & \multicolumn{1}{c|}{2.1} & 13.6 \\
RONO & CVPR'23 & \underline{90.7} & \multicolumn{1}{c|}{\underline{91.7}} & \underline{90.1} & \multicolumn{1}{c|}{\underline{90.6}} & \underline{88.8} & \multicolumn{1}{c|}{\underline{89.4}} & \underline{87.6} & \multicolumn{1}{c|}{\underline{87.4}} & \underline{89.5} & 44.4 & \multicolumn{1}{c|}{44.4} & 41.9 & \multicolumn{1}{c|}{42.1} & 38.2 & \multicolumn{1}{c|}{37.4} & 34.5 & \multicolumn{1}{c|}{32.6} & 39.4 \\
GNN4CMR & TPAMI'23 & 85.4 & \multicolumn{1}{c|}{85.1} & 80.6 & \multicolumn{1}{c|}{80.5} & 69.5 & \multicolumn{1}{c|}{68.6} & 43.9 & \multicolumn{1}{c|}{43.2} & 69.6 & 47.3 & \multicolumn{1}{c|}{47.1} & 42.0 & \multicolumn{1}{c|}{40.9} & 35.8 & \multicolumn{1}{c|}{34.4} & 29.9 & \multicolumn{1}{c|}{28.3} & 38.2 \\
DHRL & TBD'24 & 90.6 & \multicolumn{1}{c|}{91.1} & 86.6 & \multicolumn{1}{c|}{86.5} & 73.5 & \multicolumn{1}{c|}{73.3} & 43.8 & \multicolumn{1}{c|}{45.5} & 73.9 & 37.1 & \multicolumn{1}{c|}{37.3} & 24.3 & \multicolumn{1}{c|}{24.1} & 14.0 & \multicolumn{1}{c|}{13.7} & 6.2 & \multicolumn{1}{c|}{7.1} & 20.5 \\
NRCH & ACM MM'24 & 88.4 & \multicolumn{1}{c|}{89.8} & 89.7 & \multicolumn{1}{c|}{90.6} & 85.5 & \multicolumn{1}{c|}{86.1} & 84.8 & \multicolumn{1}{c|}{85.9} & 87.6 & 42.3 & \multicolumn{1}{c|}{43.6} & 42.1 & \multicolumn{1}{c|}{43.4} & 41.1 & \multicolumn{1}{c|}{42.6} & 40.6 & \multicolumn{1}{c|}{41.9} & 42.2 \\
DRCL & TMM'25 & 83.2 & \multicolumn{1}{c|}{81.5} & 64.0 & \multicolumn{1}{c|}{64.2} & 35.8 & \multicolumn{1}{c|}{37.1} & 12.2 & \multicolumn{1}{c|}{13.4} & 48.9 & 47.2 & \multicolumn{1}{c|}{46.5} & 34.7 & \multicolumn{1}{c|}{32.4} & 17.4 & \multicolumn{1}{c|}{16.2} & 5.5 & \multicolumn{1}{c|}{5.1} & 25.6 \\
RSHNL & AAAI'25 & \underline{91.2} & \multicolumn{1}{c|}{\underline{91.2}} & \underline{90.2} & \multicolumn{1}{c|}{\underline{90.0}} & \underline{87.8} & \multicolumn{1}{c|}{\underline{87.1}} & \underline{86.2} & \multicolumn{1}{c|}{\underline{84.9}} & \underline{88.6} & \underline{52.7} & \multicolumn{1}{c|}{\underline{53.4}} & \underline{52.0} & \multicolumn{1}{c|}{\underline{52.3}} & \underline{49.8} & \multicolumn{1}{c|}{\underline{50.4}} & \underline{42.7} & \multicolumn{1}{c|}{\underline{43.0}} & \underline{49.5} \\
NIRNL & ours & \textbf{92.5} & \multicolumn{1}{c|}{\textbf{92.0}} & \textbf{92.1} & \multicolumn{1}{c|}{\textbf{91.6}} & \textbf{91.6} & \multicolumn{1}{c|}{\textbf{92.0}} & \textbf{91.3} & \multicolumn{1}{c|}{\textbf{91.2}} & \textbf{91.8} & \textbf{52.9} & \multicolumn{1}{c|}{\textbf{53.3}} & \textbf{52.2} & \multicolumn{1}{c|}{\textbf{52.8}} & \textbf{51.6} & \multicolumn{1}{c|}{\textbf{52.5}} & \textbf{49.9} & \multicolumn{1}{c|}{\textbf{50.8}} & \textbf{52.0} \\ \hline
\end{tabular}
\caption{The MAP scores with different noise rates on the XMedia and INRIA-Websearch datasets.}
\label{tab:2}
\end{table*}

\subsection{Dataset}
To verify the effectiveness of our NIRNL, we conduct extensive experiments on three available benchmark datasets, including Wikipedia \cite{wiki}, XMedia \cite{Xmedia}, and INRIA-Websearch \cite{INRIA-Websearch}. A detailed description of all datasets is provided as follows:
\begin{itemize}
    \item \textbf{Wikipedia} is made of 2,866 image-text pairs, which are labeled with one of 10 semantic categories. In our experiment, the image modality is represented by a 4,096-dimensional vector extracted by the pre-trained VGG-19 \cite{VGG-19} model, while the text modality is denoted as a 300-dimensional vector obtained by the pre-trained Doc2vec \cite{Doc2Vec} model. Following previous works \cite{pu2025deep}, we randomly partition it into three subsets: 2,173, 231, and 462 pairs for training, validation, and testing, respectively.
    \item \textbf{XMedia} is a benchmark dataset with 5 modalities belonging to 20 semantic labels. In this work, we just use 5,000 image-text pairs to conduct experiments. Specifically, the image modality is encoded as 4,096-dimensional vectors using a pre-trained VGG-19 model, while the text modality is represented by a 3,000-dimensional Bag-of-Words (BOW) embedding. Following previous works \cite{pu2025deep}, we randomly split it into three subsets: 4,000, 500, and 500 pairs for training, validation, and testing, respectively.
    \item \textbf{INRIA-Websearch} is a widely-used dataset that comprises over 70,000 image-text pairs from 353 semantic categories. In this work, we use a subset including 14,698 pairs within semantic categories to perform experiments. Specifically, we employ the 4096-dimensional features extracted by the pre-trained AlexNet \cite{Alex} as image input, while utilizing the 1000-dimensional features mined by the LDA \cite{hu2021learning} as text inputs. Like previous works \cite{pu2025deep}, we randomly divide it into three subsets: 9,000, 1,332, and 4,366 pairs for training, validation, and testing, respectively.
\end{itemize}

\subsection{Baselines}
To demonstrate the effectiveness and superiority of the proposed NIRNL, we compare it with 10 state-of-the-art (SOTA) cross-modal retrieval (CMR) methods, including three unsupervised methods (i.e., DGCPN \cite{DGCPN}, CIRH \cite{CIRH}, and UCCH \cite{hu2022unsupervised}), four robust methods against noisy labels (i.e., RONO \cite{feng2023rono}, DHRL \cite{DHRL}, NRCH \cite{wang2024robust}, and RSHNL \cite{RSHNL}), and three supervised methods (i.e., ALGCN \cite{ALGCN}, GNN4CMR \cite{qian2022integrating}, and DRCL \cite{pu2025deep}). For the sake of rigor, all comparisons are conducted using results reproduced under the same dataset settings.

\begin{figure*}[!h]
\centering
\subfigure[Wikipedia (I2T)]{\includegraphics[scale=0.23]{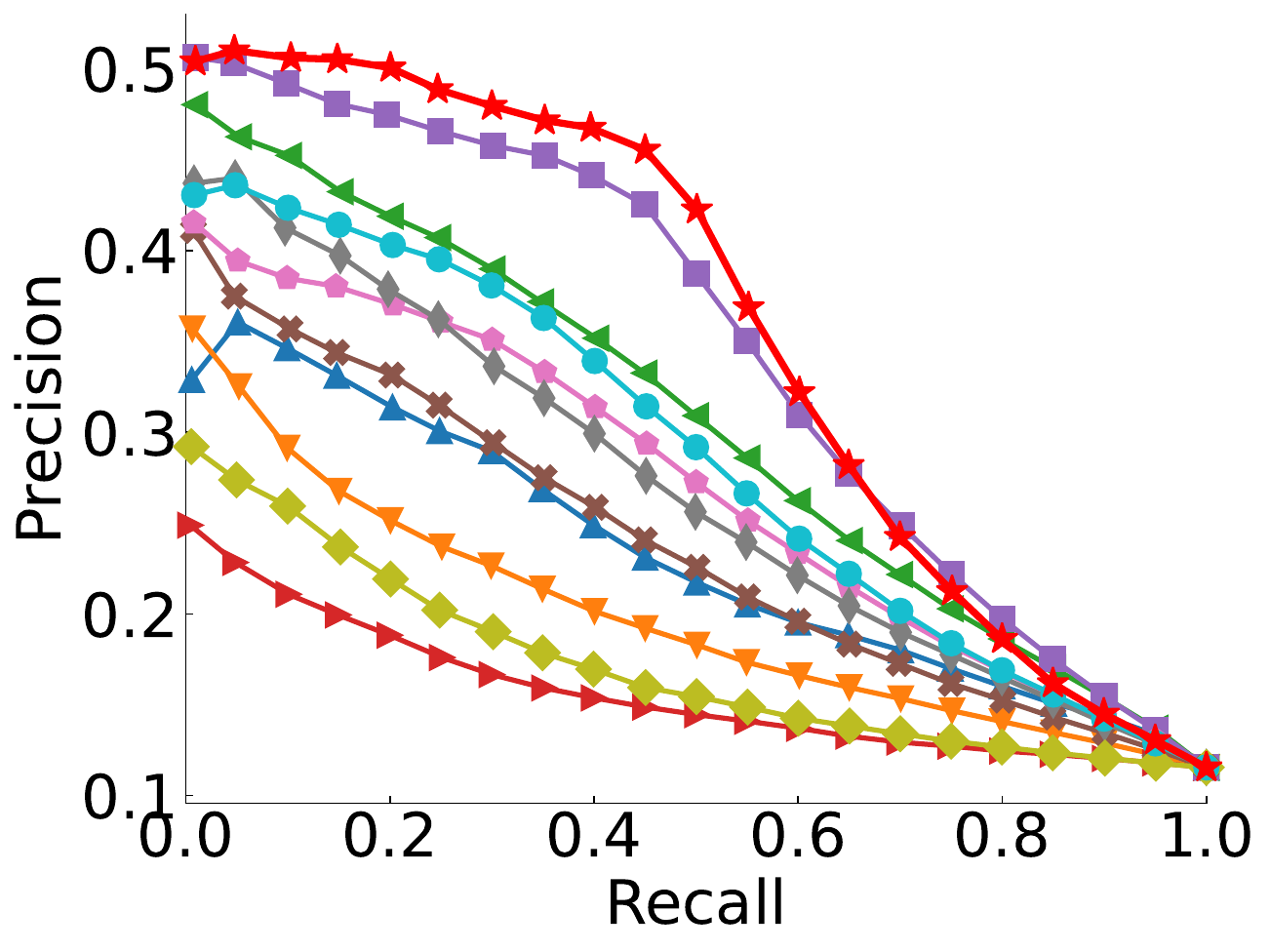}}
\subfigure[XMedia (I2T)]{\includegraphics[scale=0.23]{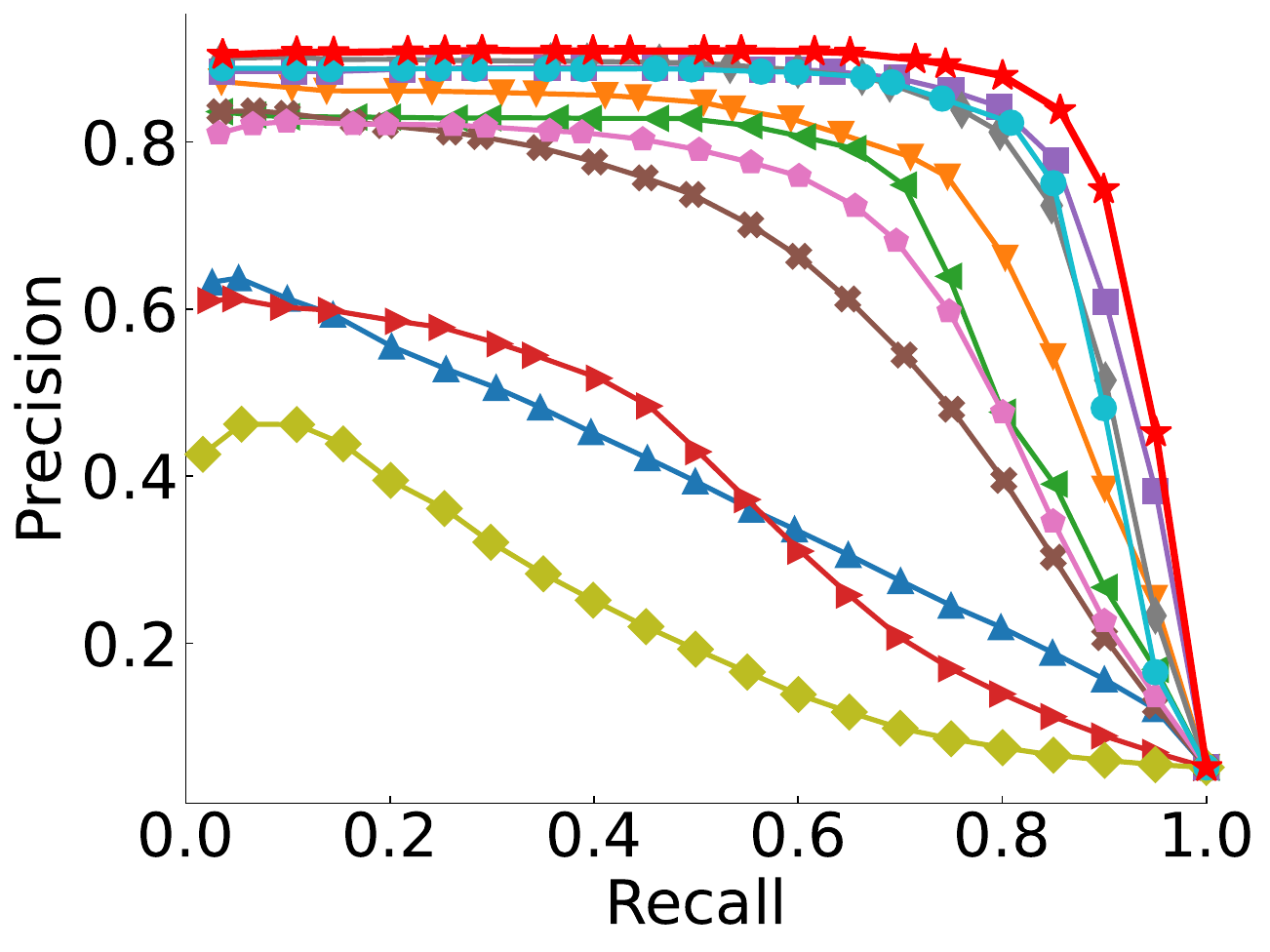}}
\subfigure[INRIA-Websearch (I2T)]{\includegraphics[scale=0.23]{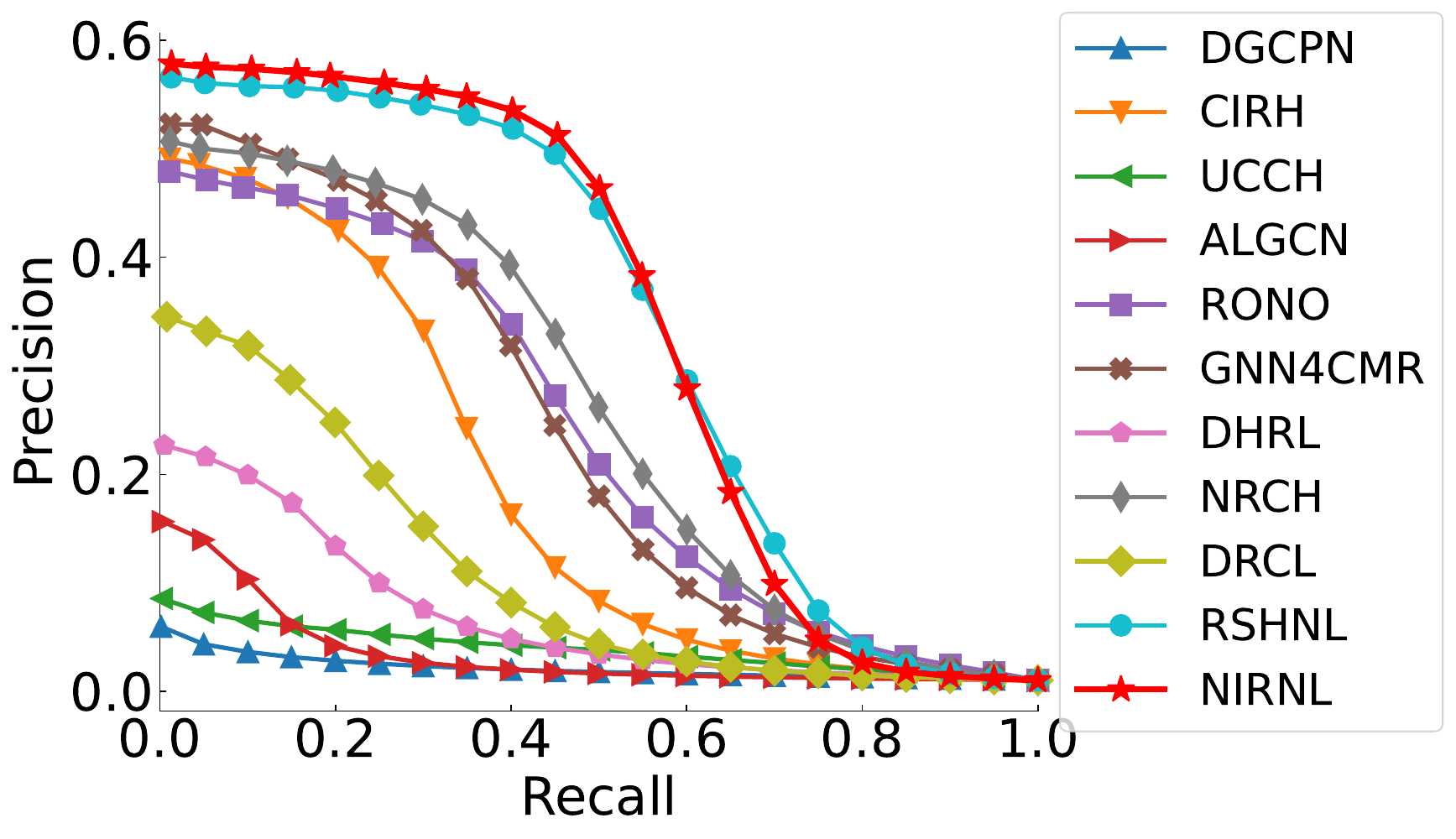}}
\\
\subfigure[Wikipedia (T2I)]{\includegraphics[scale=0.23]{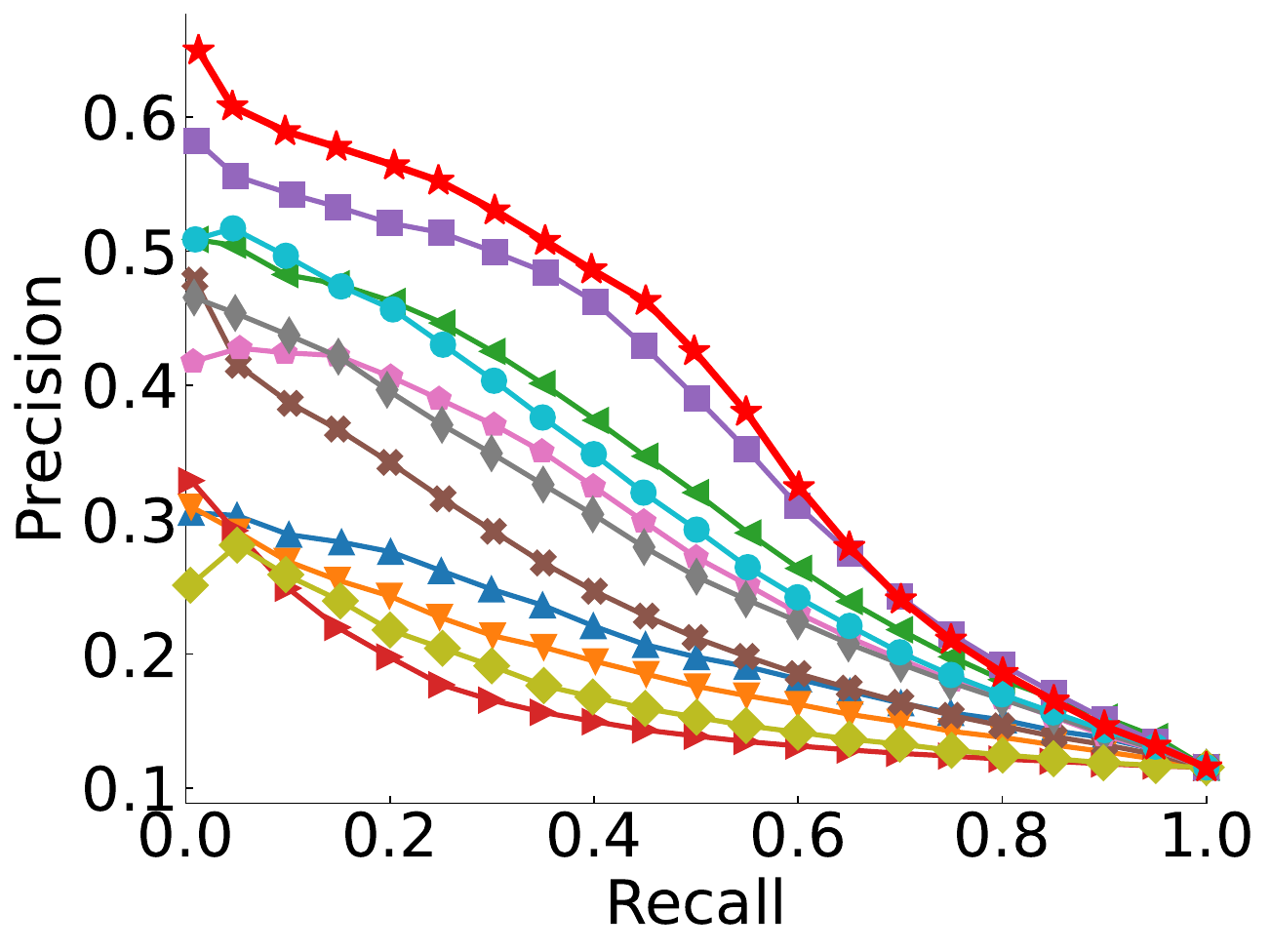}}
\subfigure[XMedia (T2I)]{\includegraphics[scale=0.23]{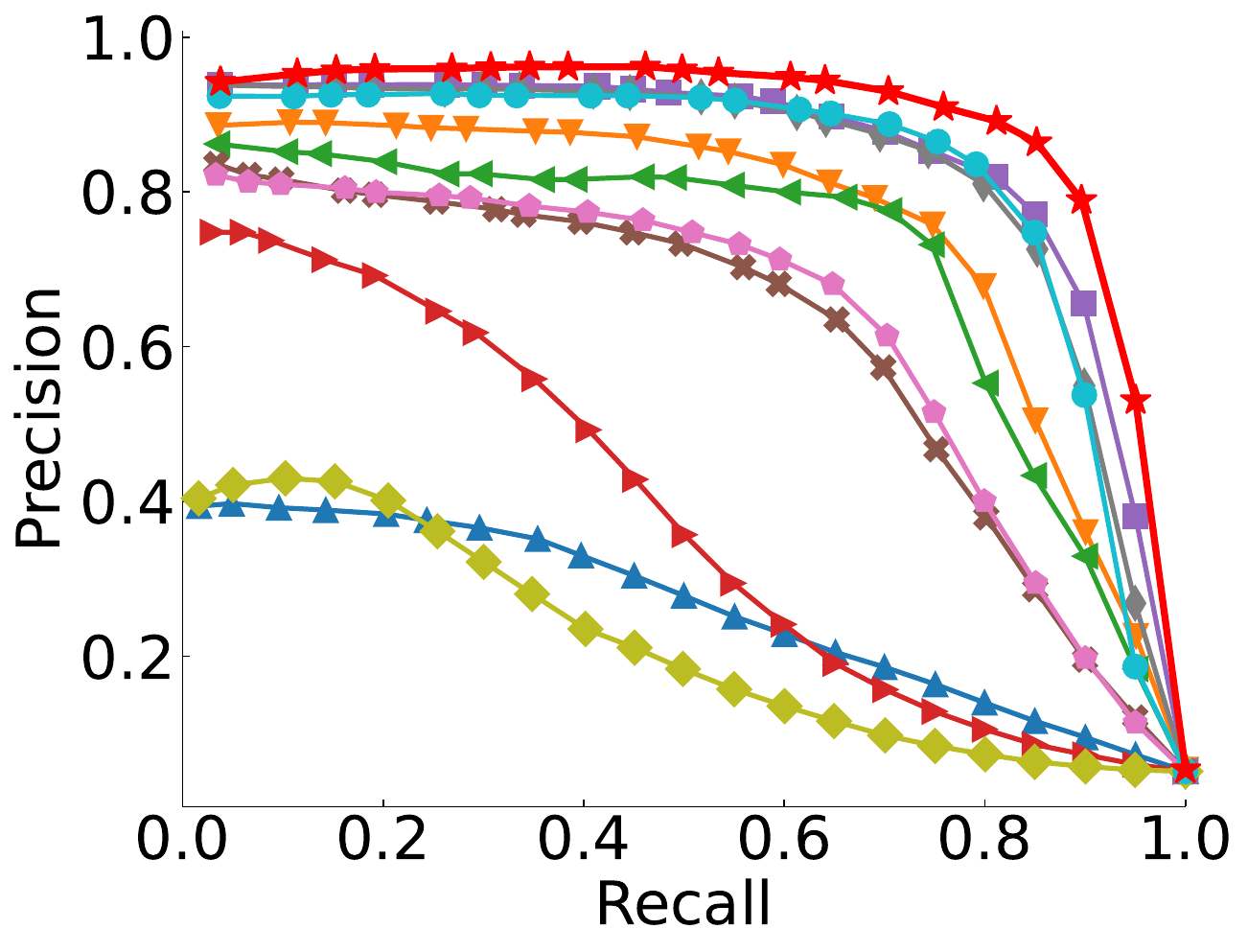}}
\subfigure[INRIA-Websearch (T2I)]{\includegraphics[scale=0.23]{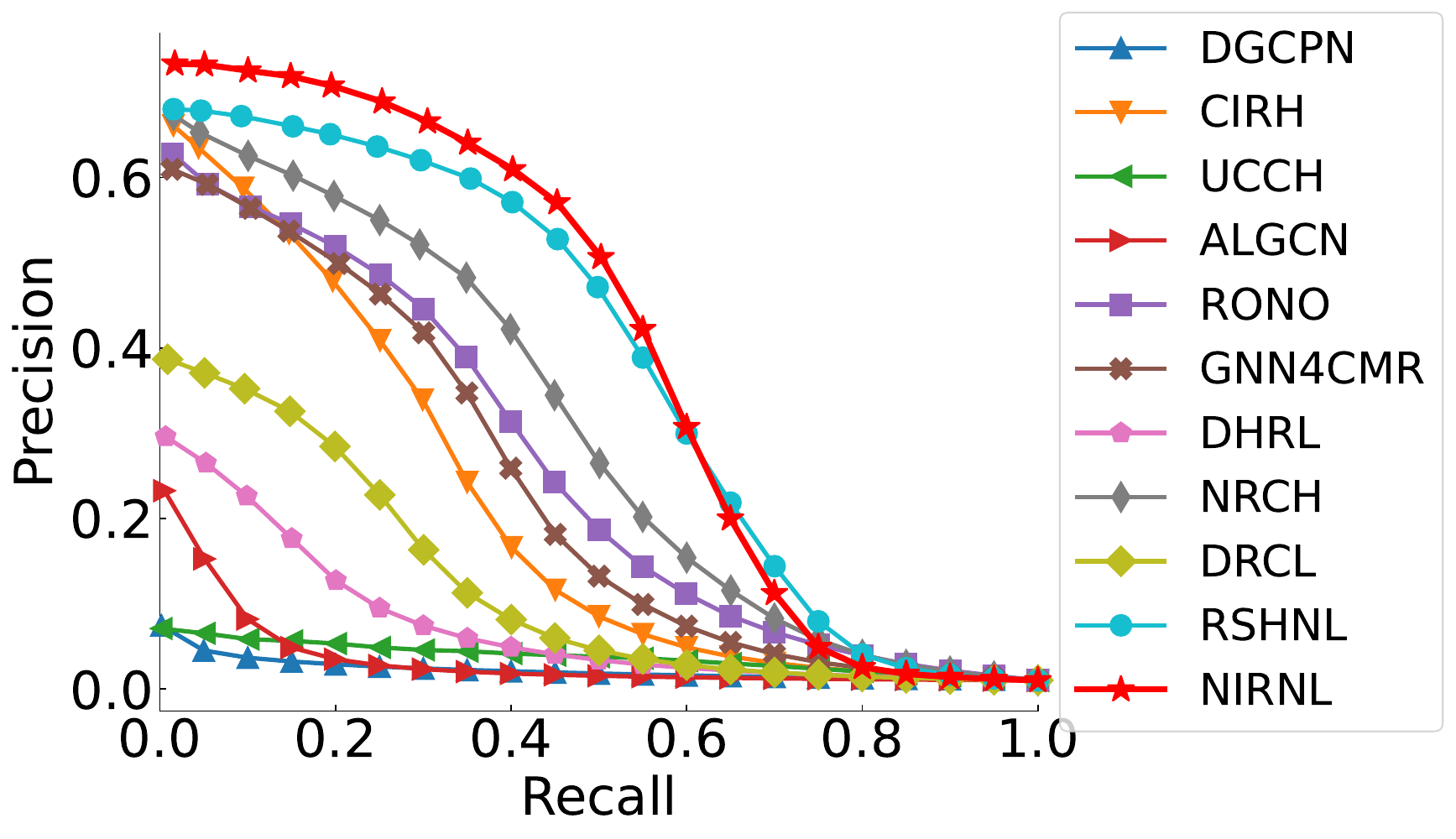}}

\caption{Precision-recall curves under the 0.6 noise rate.}
\label{fig:2}
\end{figure*}

\subsection{Experimental settings and Evaluation Metric}
To evaluate the performance of the proposed NIRNL and SOTA methods, we perform two common CMR tasks, i.e., using image modality samples as queries to retrieve semantically similar samples from the text modality (I2T), and using text modality samples as queries to retrieve semantically similar samples from the image modality (T2I). Following previous work \cite{RSHNL}, we assess the robustness of each method under varying levels of symmetric label noise, with noise rates set to 0.2, 0.4, 0.6, and 0.8, respectively. To intuitively quantify retrieval performance, we adopt the Mean Average Precision (MAP) score \cite{ALGCN, pu2025deep} as the evaluation metric. For fairness, we keep the original backbone networks frozen during training and report the MAP scores on the test set at the epoch where MAP reaches its peak on the validation set.

\begin{figure*}[h]
\centering
\subfigure[Wikipedia ]{\includegraphics[scale=0.235]{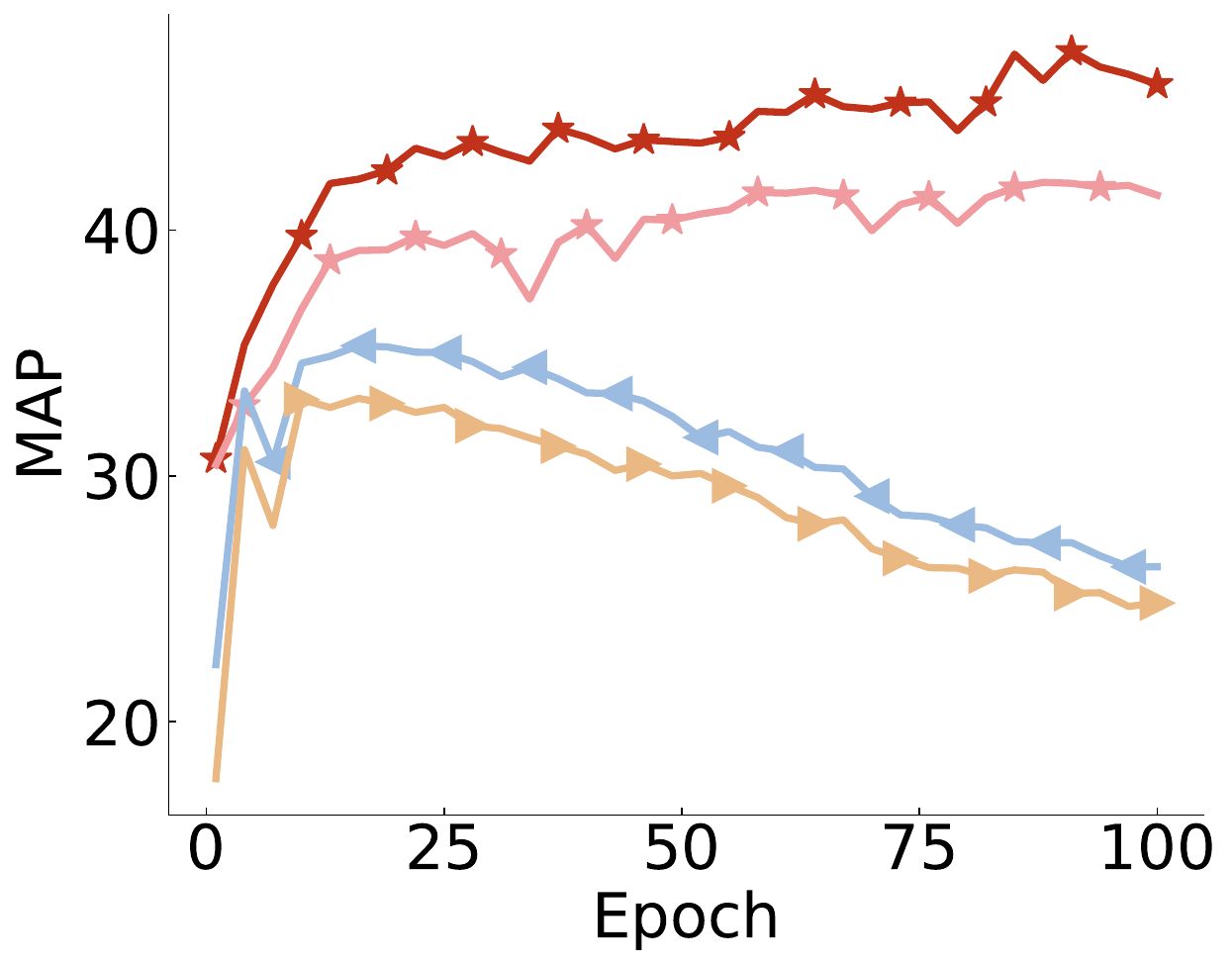}}
\subfigure[XMedia ]{\includegraphics[scale=0.235]{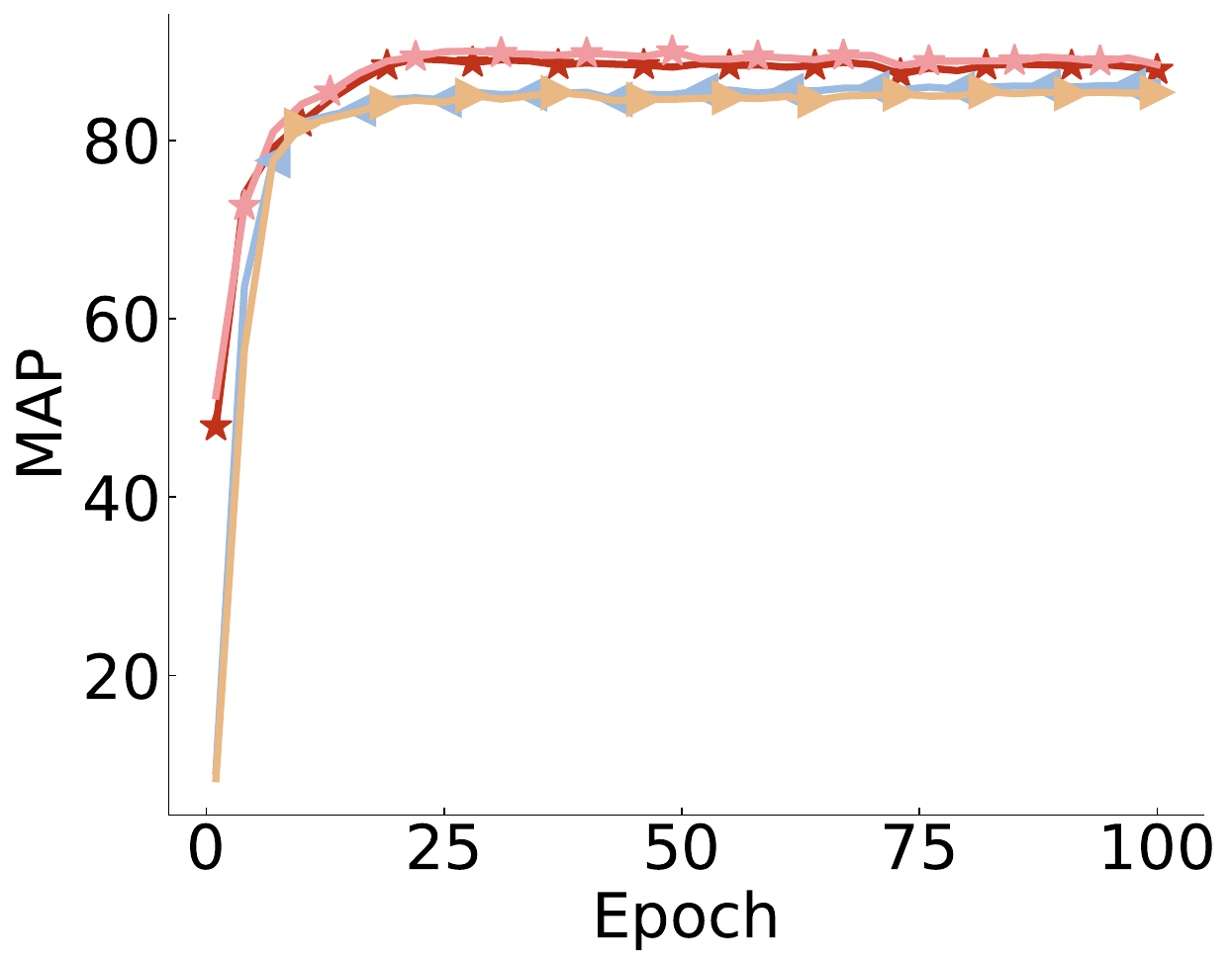}}
\subfigure[INRIA-Websearch ]{\includegraphics[scale=0.235]{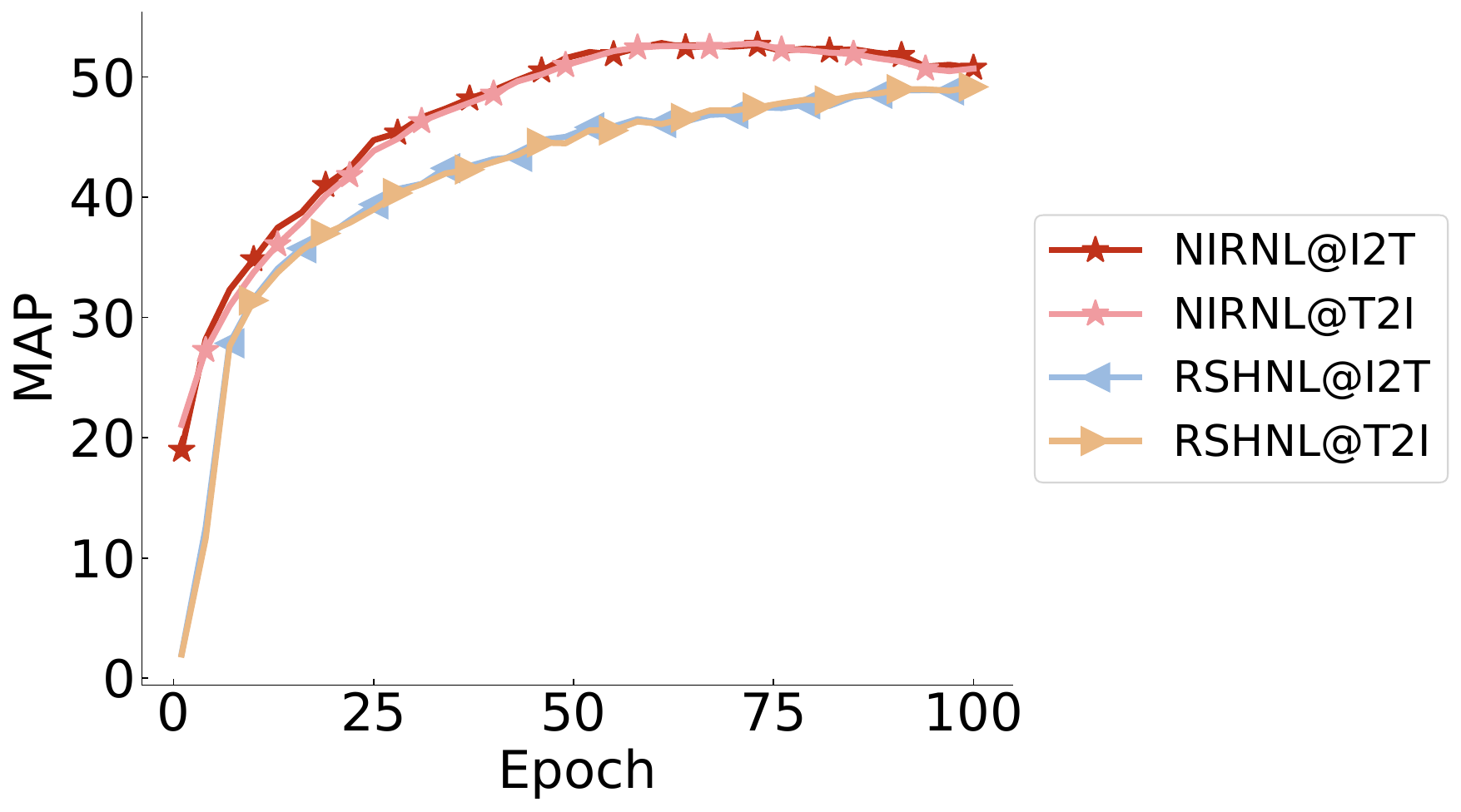}}

\caption{The MAP scores versus epochs under the 0.6 noise rate.}
\label{fig:3}
\end{figure*}

\begin{figure*}[h]
\centering
\subfigure[Wikipedia ]{\includegraphics[scale=0.195]{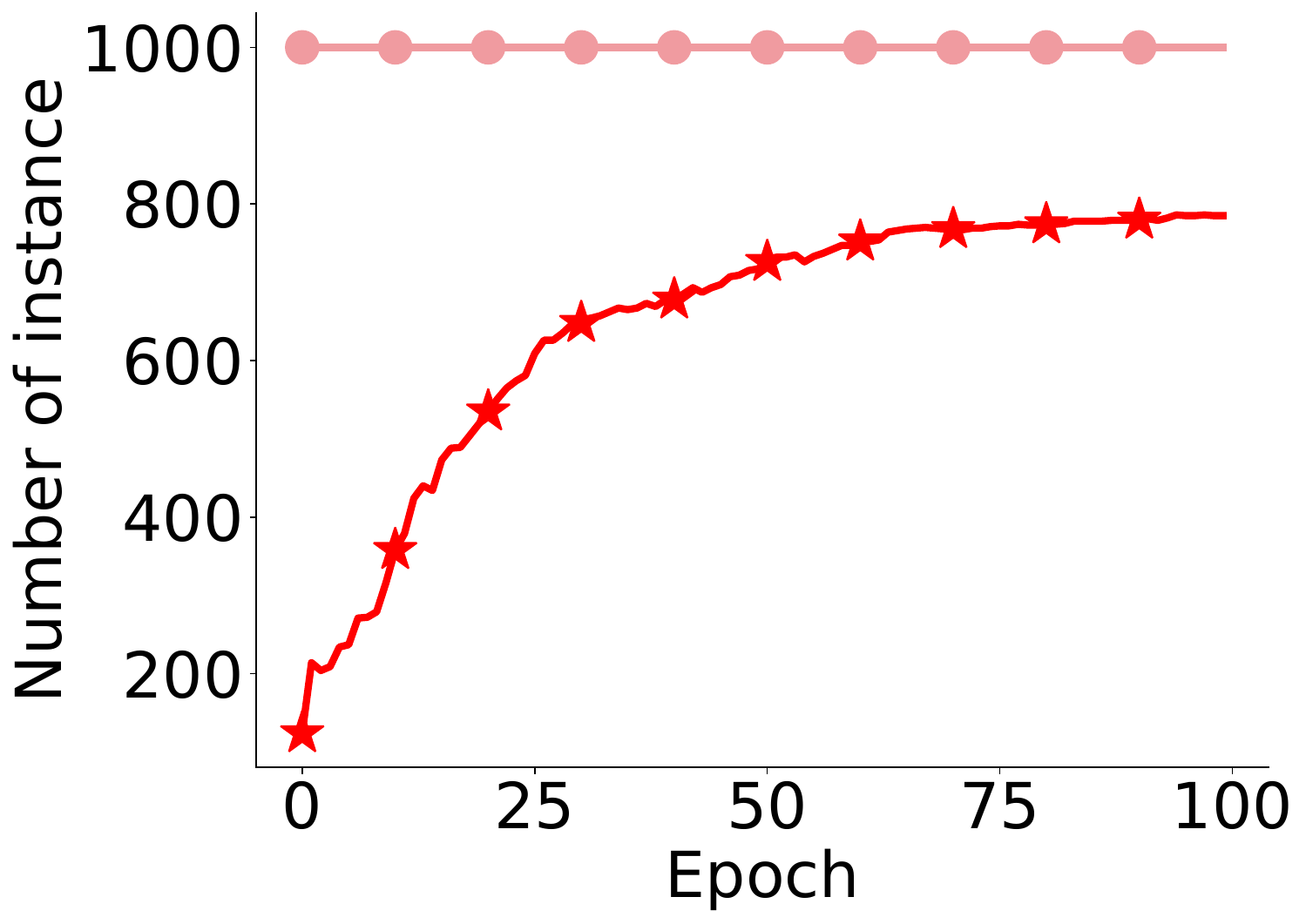}}
\subfigure[XMedia ]{\includegraphics[scale=0.195]{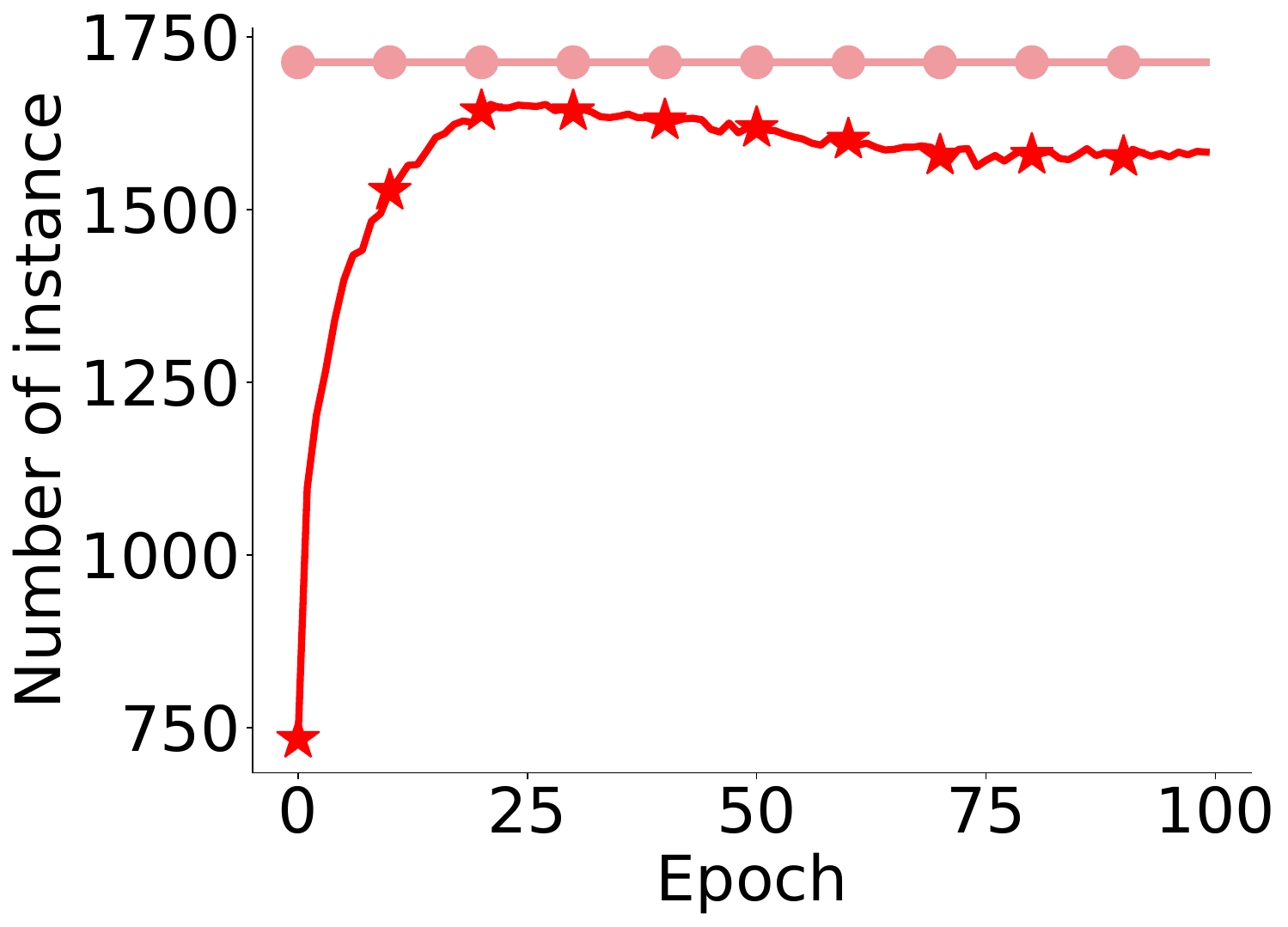}}
\subfigure[INRIA-Websearch ]{\includegraphics[scale=0.195]{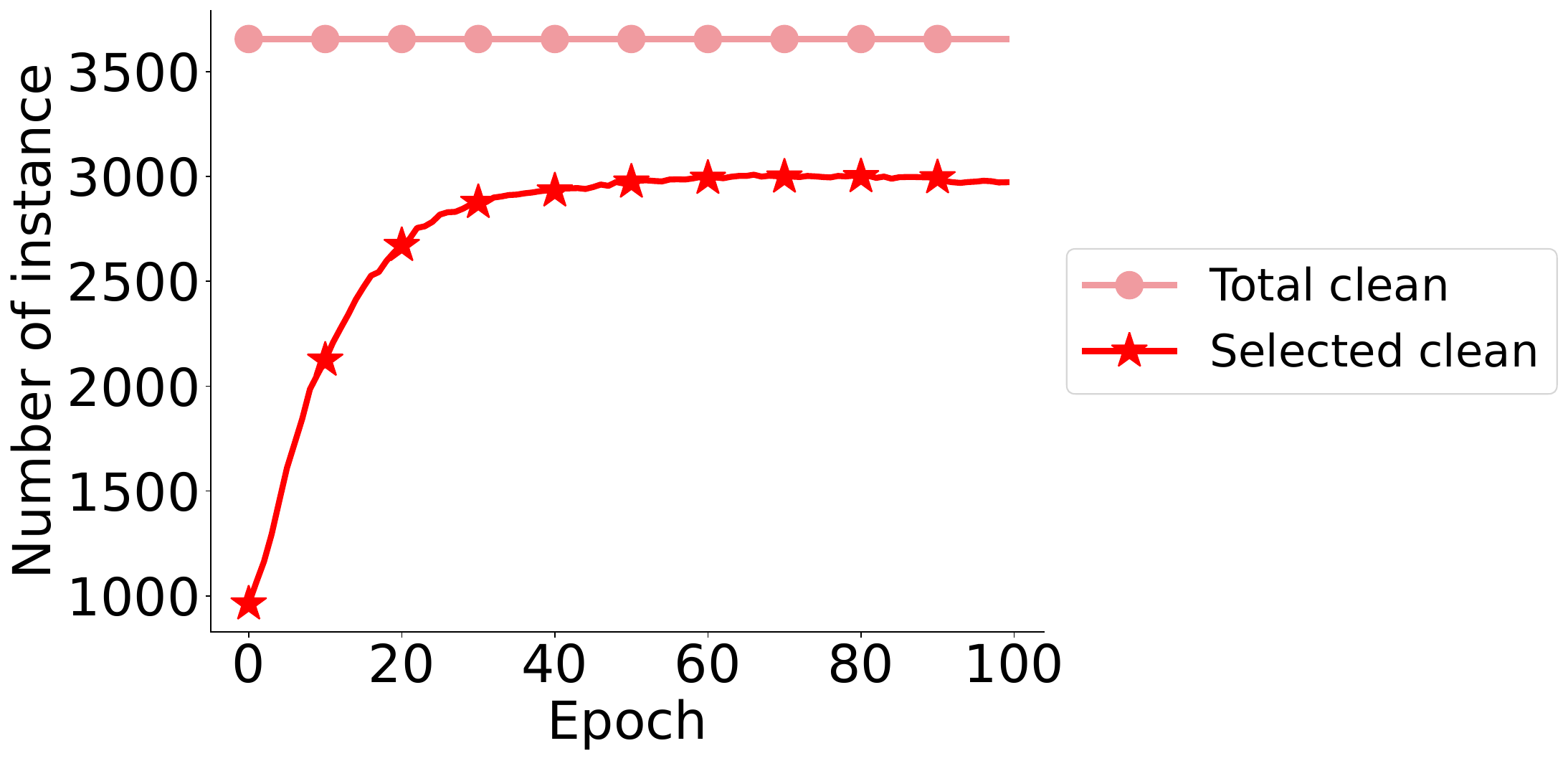}}

\caption{The number of instances versus epochs under the 0.6 noise rate.}
\label{fig:4}
\end{figure*}

\subsection{Comparisons with State-of-the-Art Methods}
To verify the superiority of NIRNL, we compare it with 10 SOTA methods in terms of MAP scores. Tab.\ref{tab:1} and Tab.\ref{tab:2} report the MAP scores on the three datasets under varying noise rates. In the tables, the second-best scores are \underline{underlined}, and the best scores are highlighted in \textbf{bold}. Then, we provide the precision-recall curves under the 0.6 noise rate in Fig.\ref{fig:2} to further assess the retrieval performance of NIRNL. 
According to the detailed analysis of these results, we can draw the following conclusion:
\begin{itemize}
    \item Although unsupervised methods (i.e., DGCPN, CIRH, and UCCH) are inherently unaffected by label noise due to the absence of supervision signals, their performance is limited by a clear bottleneck, as evidenced by their inferior results compared to some supervised methods (such as RONO and NIRNL).
    \item Some supervised methods (i.e., ALGCN, GNN4CMR, and DRCL) heavily rely on accurate annotations. As the noise rate increases, they suffer significant performance degradation or even fail due to their inability to handle noisy labels. In contrast, some robust supervised methods (i.e., RONO, NRCH, RSHNL, and NIRNL) demonstrate greater robustness by incorporating mechanisms specifically designed to mitigate the impact of label noise.
    \item From the PR-curves, it shows that the curve of NIRNL lies above that of other methods, demonstrating NIRNL can achieve higher retrieval precision at the same recall rates. In general, by refining training instances to capture the global neighborhood structure and exploit all potential information, our proposed NIRNL demonstrates superior and more robust performance.
\end{itemize}

\subsection{Ablation Study}
To investigate the contribution of specific components in our method, we conduct ablation experiments on the three datasets at a 0.6 noisy rate. Specifically, we construct four variants and the corresponding results are reported in Tab.\ref{tab:3}, where \textbf{NIRNL-1} denotes the removal of CMP, \textbf{NIRNL-2} represents the discarding of the noisy subset, \textbf{NIRNL-3} means treating the hard subset the same as the pure subset, without applying instance loss-weighting, and \textbf{NIRNL-4} refers to the exclusion of the barycenters alignment mechanism for all samples and only using CMP to optimize the model. From the results, it could be concluded that: 
\begin{itemize}
    \item The removal of CMP weakens the discriminability among inter-class samples, thereby leading to a drop in retrieval performance. 
    \item The noisy subset contains potentially useful information, which can be leveraged through label correction to improve overall performance. In contrast, directly discarding these samples may lead to suboptimal results due to the loss of informative signals. 
    \item When the loss-weighting mechanism designed for the hard subset is deleted, the model treats all samples as equally reliable, which amplifies the effect of noisy labels and results in performance degradation.
    \item Excluding the barycenter alignment mechanism is equivalent to discarding label information as a supervisory signal, leading to a noticeable decline in performance. 
\end{itemize}
Overall, the ablation studies highlight the complementary roles of each component within the NIRNL framework.

\begin{table}[h]
\centering
\setlength{\tabcolsep}{0.8pt}
\begin{tabular}{c|ccc|ccc|ccc}
\hline
 Dataset& \multicolumn{3}{c|}{Wikipedia} & \multicolumn{3}{c|}{XMedia} & \multicolumn{3}{c}{Websearch} \\ 
 Task& I2T & T2I & mean & I2T & T2I & mean & I2T & T2I & mean \\ \hline
NIRNL-1 & 25.9 & 22.9 & 24.4 & 40.0 & 41.6 & 40.8 & 7.9 & 8.7 & 8.3 \\ 
NIRNL-2 & 47.1 & 42.4 & 44.8 & 88.8 & 88.4 & 88.6 & 46.3 & 47.1 & 46.7 \\ 
NIRNL-3 & 48.4 & 45.8 & 47.1 & 90.0 & 90.6 & 90.3 & 50.5 & 51.6 & 51.1 \\ 
NIRNL-4 & 41.2 & 39.8 & 40.5 & 90.1 & 91.4 & 90.8 & 49.9 & 50.9 & 50.4 \\ 
NIRNL & \textbf{49.2} & \textbf{46.1} & \textbf{47.7} & \textbf{91.6} & \textbf{92.0} & \textbf{91.8} & \textbf{51.6} & \textbf{52.5} & \textbf{52.1} \\ \hline
\end{tabular}
\caption{Ablation study under the 0.6 noise rate, where `Websearch' denotes the INRIA-Websearch dataset.}
\label{tab:3}
\end{table}

\subsection{Robustness Analysis}
To intuitively demonstrate the robustness of the proposed NIRNL, we compare it with RSHNL in terms of MAP scores versus epochs under the 0.6 noise rate. As illustrated in Fig.\ref{fig:3}, our NIRNL consistently outperforms RSHNL across all three datasets. Specifically, RSHNL exhibits lower performance on the XMedia and INRIA-Websearch datasets due to its failure to leverage noisy samples. On the Wikipedia dataset, RSHNL even shows signs of overfitting to noisy labels, as it fails to capture the global distributional structure of neighboring instances. In contrast, NIRNL employs the Neighbor-aware Instance Refining (NIR) module to adaptively refine all instances by capturing global distributional information, thereby maximizing the use of available sample information and achieving superior performance.

Additionally, to further evaluate the robustness of NIRNL, we track the number of three types of instances across different training epochs under a 0.6 noise rate, including the total number of clean instances in the training set and the number of truly clean instances within this selected pure subset. As shown in Fig.\ref{fig:4}, NIRNL can incorporate the majority of clean samples into the training process over time. This may be attributed to the Neighbor-aware Instance Refining (NIR) mechanism, which enables the model to perceive the global distribution of instances and refine instances to maximize the use of all available information.
These findings strongly support the effectiveness and robustness of the proposed NIRNL.

\section{Conclusion} 
In this paper, to robustly learn with noisy labels, we propose a novel unified cross-modal learning framework, termed Neighbor-aware Instance Refining with Noisy Label (NIRNL). Our NIRNL inherits the advantages of the three existing paradigms to balance the upper limit of model performance, calibration reliability, and data utilization rate. Specifically, we first construct Cross-modal Margin Preserving (CMP) to structurally regularize the embedding space to boost representation discriminability. Then, we propose Neighbor-aware Instance Refining (NIR) to fully explore the semantic information of each instance. To be specific, NIR employs neighborhood consensus to tri-split data, thereby obtaining pure, hard, and noisy instances, respectively. Moreover, NIR designs three different loss functions to process these instances respectively, thereby achieving more robust learning with noisy labels. We conduct various experiments on three benchmarks with different noise rates. Experimental results show that our NIRNL obtains the best retrieval performance and robustness, especially under high noise rates.


\section{Acknowledgments}
This work is supported by the Central Government's Guide to Local Science and Technology Development Fund under Grant 2025ZYDF101.






\bigskip
\noindent

\bibliography{aaai2026}

\end{document}